\documentclass[sigconf]{acmart}

\graphicspath{{./fig/}}
\usepackage{multirow}

\usepackage{enumitem}
\usepackage{xcolor}
\usepackage{fixltx2e}
\usepackage{amsmath,bm}

\usepackage{soul}
\soulregister\cite7 
\soulregister\ref7

\usepackage{acmart-taps}

\AtBeginDocument{%
  \providecommand\BibTeX{{%
    \normalfont B\kern-0.5em{\scshape i\kern-0.25em b}\kern-0.8em\TeX}}}

\setcopyright{acmcopyright}
\copyrightyear{2023}
\acmYear{2023}
\acmDOI{3581641.3584048}

\acmConference[IUI 2023]{28th Annual Conference on Intelligent User Interfaces}{Mar 27--31,
  2023}{Sydney, Australia}
\acmBooktitle{IUI '23: ACM Conference on Intelligent User Interfaces,
 Mar 27--31, 2023, Sydney, Australia} 
\acmPrice{15.00}
\acmISBN{979-8-4007-0106-1/23/03}

\begin{document}

\newcommand{\hlc}[2][yellow]{{\sethlcolor{#1}\hl{#2}}}
\definecolor{revision_color}{HTML}{FFFFFF}
\definecolor{re_emphasis_color}{HTML}{FFFFFF}
\newcommand{\rev}[1]{\hlc[revision_color]{#1}} 
\newcommand{\rep}[1]{\hlc[re_emphasis_color]{#1}}

\title[IRIS: Interpretable Rubric-Informed Segmentation for Action Quality Assessment]{IRIS: Interpretable Rubric-Informed Segmentation for \\Action Quality Assessment}

\author{Hitoshi Matsuyama}
\email{hitoshi@ucl.nuee.nagoya-u.ac.jp}
\author{Nobuo Kawaguchi}
\email{kawaguti@nagoya-u.jp}
\affiliation{%
  \institution{Nagoya University}
  \country{Japan}
}

\author{Brian Y. Lim}
\authornote{Corresponding author.}
\email{brianlim@comp.nus.edu.sg}
\affiliation{%
  \institution{National University of Singapore}
  \country{Singapore}
}

\renewcommand{\shortauthors}{Matsuyama, Kawaguchi, and Lim}

\begin{abstract}
  AI-driven Action Quality Assessment (AQA) of sports videos can mimic Olympic judges to help score performances as a second opinion or for training.
  However, these AI methods are uninterpretable and do not justify their scores, which is important for algorithmic accountability.
  Indeed, to account for their decisions, instead of scoring subjectively, sports judges use a consistent set of criteria --- rubric --- on multiple actions in each performance sequence. 
  Therefore, we propose IRIS to perform Interpretable Rubric-Informed Segmentation on action sequences for AQA.
  We investigated IRIS for scoring videos of figure skating performance. 
  IRIS predicts (1) action segments, (2) technical element score differences of each segment relative to base scores, (3) multiple program component scores, and (4) the summed final score.
  In a modeling study, we found that IRIS performs better than non-interpretable, state-of-the-art models. 
  In a formative user study, practicing figure skaters agreed with the rubric-informed explanations, found them useful, and trusted AI judgments more.
  This work highlights the importance of using judgment rubrics to account for AI decisions.

\end{abstract}




\maketitle

\section{Introduction}
The need for explainable AI (XAI) has grown significantly due to the prevalence of AI in many aspects of society. This is especially important where judgements are performed automatically.
One such area is sports analytics or action quality assessment (AQA), where many AI-based computer vision techniques have been developed to assess the quality of performances from videos~\cite{li2019manipulation,li2018end,bertasius2017baller,parmar2019action,cvpr_uncertainty,iccv_groupaware,parmar_olympic,nekoui2021eagle,8756030,zeng2020hybrid}. 
This can mimic Olympic judges to help score performances as a second opinion, help spectators understand how certain sports are judged, and help athletes get access to cheaper and faster feedback. 
However, these techniques mostly predict scores and do not provide more information on how the scores were derived. They are inscrutable and uninterpretable.

There are some prior work to explain predictions in AQA.
For example, Yu et al. explained scoring diving performances~\cite{yu2021group} using Grad-CAM saliency maps~\cite{selvaraju2017grad}. 
Although saliency maps can be useful in some applications~\cite{alqaraawi2020evaluating}, 
these explanations were originally designed for model debugging, and are overly technical and unlikely to be usable by practicing athletes.
One way to increase relevance is to make the explanations concept-based~\cite{kim2018interpretability,koh2020concept} or more relatable~\cite{zhang2022towards}, but these may be too simplistic and generic for skilled activities.
Several researchers have argued that AI explanations should be tailored to end users and relevant contexts~\cite{liao2020questioning,lim2009assessing,wang2019designing}.
Wang and Yin identified several important desiderata of XAI for non-experts~\cite{wang2021explanations} and Dodge et al. demonstrated the benefit of XAI on fairness decisions~\cite{dodge2019explaining}, but there has been limited work on providing XAI to domain experts~\cite{lim2023diagram, lyu2022if}.
Pirsiavash et al. developed diving feedback proposals by calculating gradients of scores relative to pose estimation joints~\cite{10.1007/978-3-319-10599-4_36}, but is only data-driven, rather than driven by how athletes or judges make decisions.

\begin{figure}[t]
  \centering
  \includegraphics[width=5.5cm]{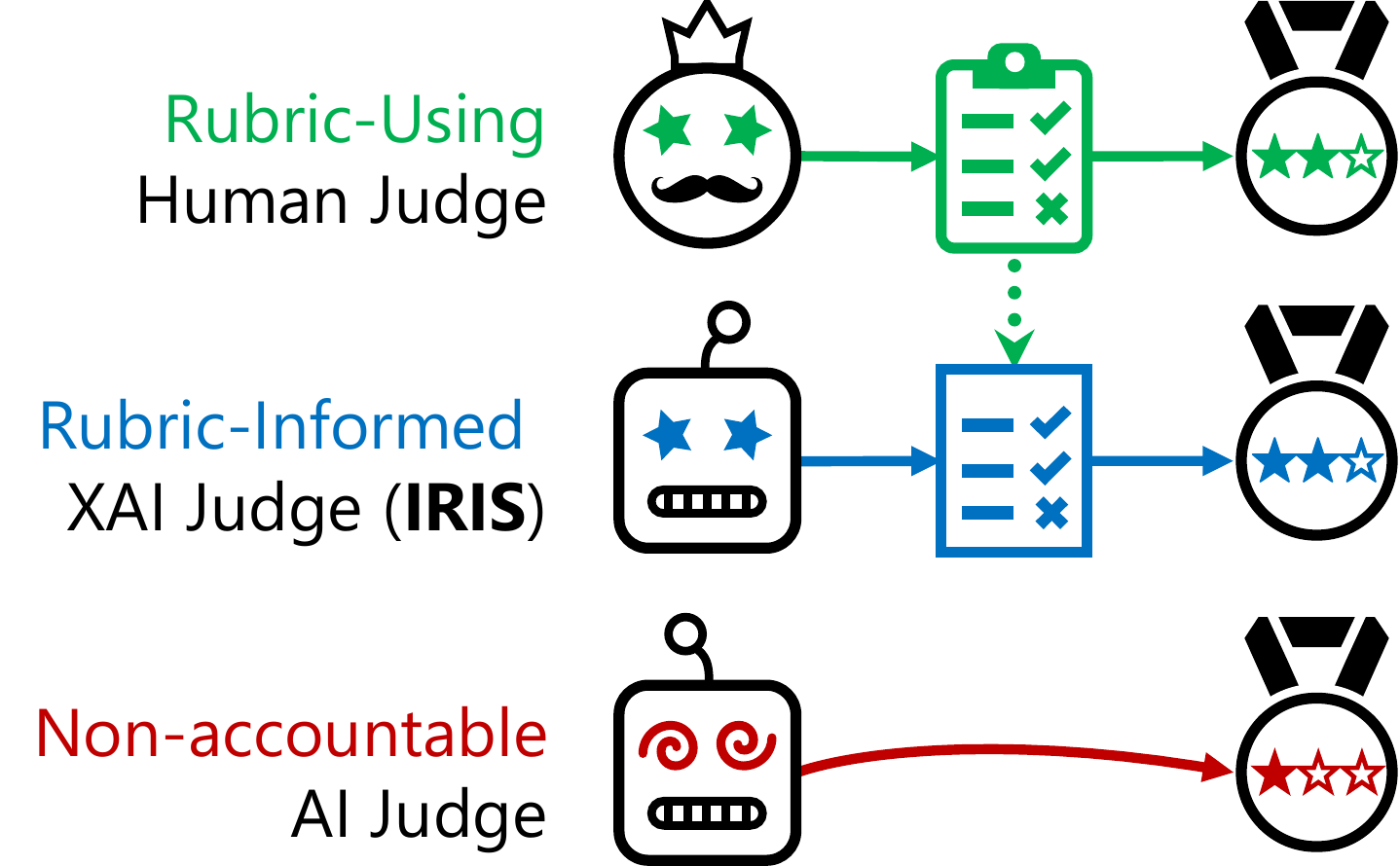}
  \caption{
  \rev{Conceptual overview of how human judges use rubrics to guide their ratings, while non-explainable AI models predict with no accountability.
  We propose rubric-informed explainable AI (XAI) judges to use rubrics from domain experts to judge with accountability.}
  }
  \label{fig:rubric-based-xai}
\end{figure}

Instead, we argue that XAI developers should study the scoring rubrics used in the application domain to identify requirements for justifications. 
Indeed, to account for their decisions, instead of scoring subjectively, sports judges use a consistent set of criteria — rubric — on multiple actions in each performance sequence. 
For example, in figure skating, judges refer to a score sheet that is standardized by the ISU Judging System~\cite{isu_rule,isu_rule_pcs} to include multiple criteria and methods for scoring.
It is important for the AI to adhere to these rubrics to be as accountable as human judges.
\rev{See Fig. \ref{fig:rubric-based-xai} for conceptual overview, and Fig. \ref{overview_ui} for demo.}

Towards this end, we propose IRIS, an \textbf{I}nterpretable \textbf{R}ubric-\textbf{I}nformed \textbf{S}egmentation method to predict and explain video-based action quality assessment (AQA).
It is interpretable to explain how it derived its judgement score.
It is informed by a rubric to determine what to consider when calculating its judgement.
It performs segmentation to locate specific sections of a performance to judge specific criteria.
We implemented this for the application domain of figure skating, but the high-level approach is generalizable to other video-based AQA (e.g., diving, dancing, cooking) with predefined criteria rubrics.
IRIS provides four key features to predict and explain its judgement: Score, Sequence, Segments, and Subscores.
Being rubric-informed, IRIS is more accurate to predict scores due to rule guidance, and is more accountable to justify its decisions based on domain-established criteria.


We evaluated IRIS in multiple studies: 
1) a demonstration study of the interpretability of IRIS with a visualization of ice skating,
2) a quantitative modeling study to compare IRIS against the baseline models, and
3) a formative user study with experienced figure skaters to qualitatively investigate the usefulness and trustworthiness of various rubric-informed features in IRIS.
In summary, our \textbf{contributions} are:
\begin{itemize}[leftmargin=*]
  \item The introduction of a \textit{rubric-informed} approach to identify interpretability requirements for AI based on scoring rubrics.
  \item The development of IRIS, an interpretable deep learning model that is more useful, understandable, and trustworthy than other state-of-the-art models for action quality assessment.
  \item An evaluation of rubric-informed interpretability for figure skating with domain experts in a formative study.
\end{itemize}



\section{Related Work}
We summarize related works on \rev{predicting and explaining AQA}.

\subsection{Action Quality Assessment (AQA)}
Many AI-driven action quality assessment (AQA) systems have been proposed for sports~\cite{li2019manipulation,li2018end,bertasius2017baller,parmar2019action,cvpr_uncertainty,iccv_groupaware,parmar_olympic,nekoui2021eagle,8756030,zeng2020hybrid} and other fields~\cite{doughty2019pros,doughty2018s,zhang2014relative}. 
AI-based AQA extracts video features trains a regression model to predict scores. 
The feature extraction methods generally use three-dimensional convolutional neural networks (3DCNN), such as C3D~\cite{c3d_arxiv} and I3D~\cite{i3d_arxiv}. These networks convert short segments of video into feature vectors by convolving in both 2D spatial and 1D temporal dimensions. 
To handle longer videos with longer-term behaviors, Parmar et al. combined the 3DCNN with recurrent LSTM to propose C3D-LSTM~\cite{parmar_olympic}.
%
Zheng et al. developed a hybrid approach of video and image features and proposed a context-aware AQA based on a Graph Convolutional Network (GCN)~\cite{zeng2020hybrid}. They modeled static posture and dynamic movement to capture the short and long-term temporal relationships. 
Nekoui et al.~\cite{nekoui2021eagle} proposed a CNN-based approach that captures both fine and coarse-grained temporal dependencies. Using both video features and pose estimation heatmaps~\cite{nekoui2020falcons}, they stacked CNN-based modules of various kernel sizes to capture patterns at different time resolutions. 
For figure skating AQA, Xu et al.~\cite{8756030} proposed a multi-scale and skip-connected CNN-LSTM method to 
capture short-term time dependencies with variously-sized CNN kernels, and LSTM with self-attention for longer-term time dependencies.

\subsection{Explaining Action Quality Assessment}
While the aforementioned research on AQA focused on high prediction performance (e.g., Spearman's rank correlation coefficient between AI and human judges), they neglected how users would use and interpret the score predictions. 
Explainable AI (XAI) techniques could help.
They are widely used in prediction tasks, such as computer vision~\cite{selvaraju2017grad,bach2015pixel,alqaraawi2020evaluating}, and natural language~\cite{millecamp2019explain,costa2018automatic,qian2021xnlp,rhys2021directed}.
For example, Yu et al. explained the scoring of diving performances~\cite{yu2021group} using Grad-CAM saliency maps~\cite{selvaraju2017grad}. 
Unfortunately, these explanations were primarily designed for model debugging, are overly technical, and unlikely to be usable by practicing athletes.
It remains unclear how practicing athletes, not data scientists, would use and understand the AI-based AQA systems.
Therefore, it is important to study how we can show or explain the AQA results to practitioners. 

Pirsiavash et al. developed feedback for the sport of diving by calculating the gradients of scores relative to pose estimation joints~\cite{10.1007/978-3-319-10599-4_36}. This is highly relevant to the diving activity, but is only data-driven, rather than driven by how athletes or judges make decisions.
Instead, we argue that XAI for AQA in sports should adhere to the rubrics that are used in the sport, which is our approach with IRIS.
Khan et al.~\cite{khan2017activity} developed a method for analyzing batting in cricket using wearable sensors and further proposed a visualization inspired by TV shows of cricket games. 
They classified various low-level sub-actions identified from careful analysis of how players move when playing cricket.
Thompson et al.~\cite{thompson2015dancing} proposed a visualization for dressage (horse riding), where they quantified scoring criteria inspired from score cards in the sport. 
We drew our rubrics from score cards in the sport of figure skating.
Both aforementioned methods used locomotion data of IMU sensors, and implemented heuristically or with shallow machine learning models (e.g., SVM, kNN, decision tree).
With IRIS, we imbued a deep learning meta-architecture with rubric-informed features to be generalizable and can apply to the more ubiquitous video data.

\section{Background: Rubric for figure skating action quality assessment}
In this section, we describe 
how rubrics are critical for formal and accountable judgement of many performances, 
our application use case of figure skating and its judging rubric.
This background guides the subsequent design of our proposed explainable system.

\begin{table*}[h]
  \centering
  \includegraphics[width=\textwidth]{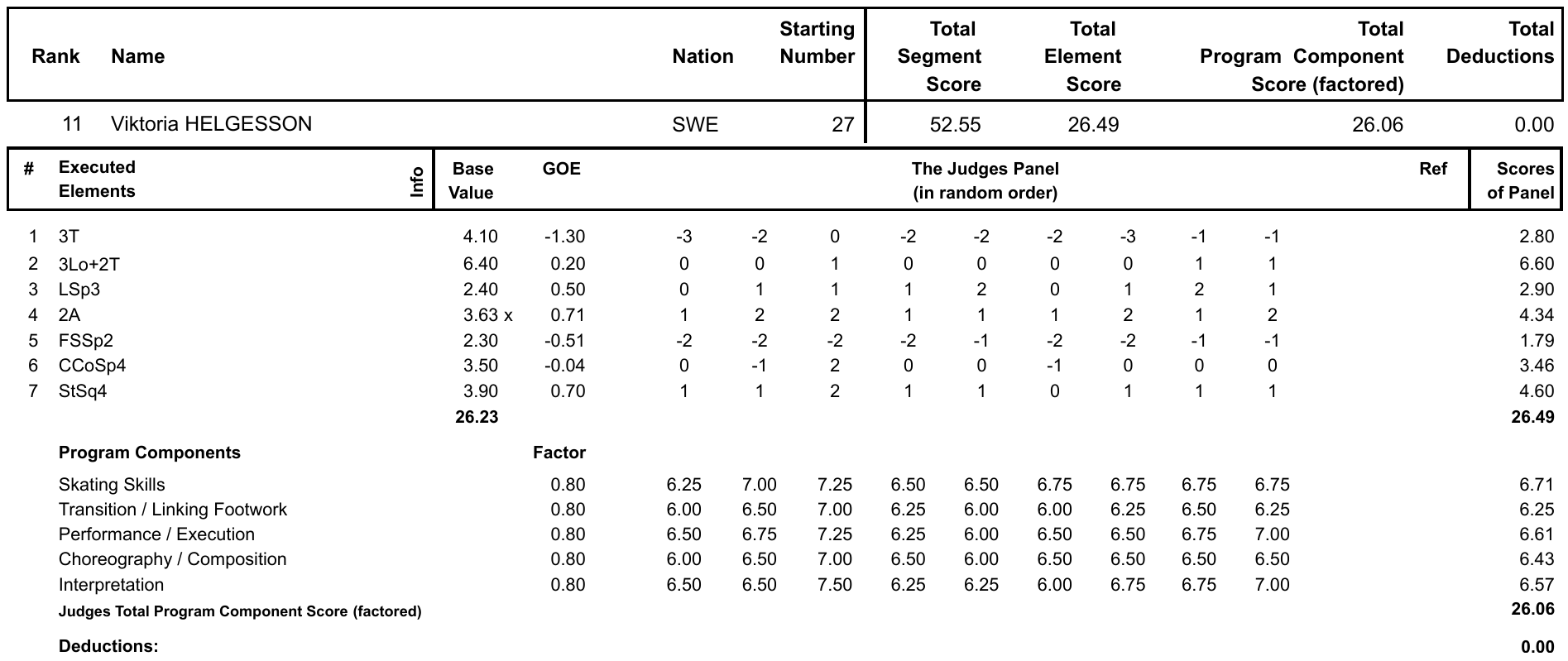}
  \vspace{0.02cm}
  \caption{
  Example score sheet showing the total score, TES, PCS, executed elements (actions that the figure skater performed), base values (difficulty scores for each action), and PCS components. 
  Judges use the score sheet to guide their judging.
  Figure skaters and spectators see the score sheets to understand the result of contests.
  }
  \label{score_sheet}
\end{table*}

\subsection{Rubrics}
In many situations where judgements affect many people's lives, scoring rubrics have been developed to formalize how performance is judged. This is especially used in education to evaluate qualitative assessments, such as essays and presentations~\cite{dawson2017assessment}, and in subjective sports~\cite{brown2008introduction}. Rubrics can help judges focus on specific aspects for evaluation, and make judging more objective, rather than based on inscrutable, subjective opinions. This also helps students and trainees know what to aim for to score highly.
Herman et al.~\cite{herman1992practical} identified four elements that make up scoring rubrics, which we clarify as two key points with corresponding elaboration points:
\aptLtoX[graphic=no,type=env]{
\begin{enumerate}[leftmargin=*]
    \item[1)] One or more traits or \textbf{dimensions} that serve as the basis for judging the student response.
    \begin{itemize}
        \item \textbf{Definitions} and \textbf{examples} to clarify the meaning of each trait or dimension.
    \end{itemize}
    \item[2)] A \textbf{scale} of values on which to rate each dimension.
    \begin{itemize}
        \item \textbf{Standards} of excellence for specified performance levels accompanied by models or examples of each level.
    \end{itemize}
\end{enumerate}
}{
\begin{enumerate}[label={\arabic*)}, leftmargin=*]
    \item One or more traits or \textbf{dimensions} that serve as the basis for judging the student response.
    \begin{itemize}
        \item \textbf{Definitions} and \textbf{examples} to clarify the meaning of each trait or dimension.
    \end{itemize}
    \item A \textbf{scale} of values on which to rate each dimension.
    \begin{itemize}
        \item \textbf{Standards} of excellence for specified performance levels accompanied by models or examples of each level.
    \end{itemize}
\end{enumerate}
}

Therefore, a rubric needs to have dimensions (also known as \textit{criteria}) on which to judge the performance, and a scale rating (or \textit{score}) for each criteria.
We focus on the high-level key points, since we are leveraging a known rubric, rather than developing them and training new judges or trainees. 
We assume that users of rubric-informed explainable AI (XAI) will have the implicit knowledge of definitions and examples, and know how to score each criteria.

\subsection{Scoring figure skating}
Figure skating is an competitive sport where athletes skate to music, to demonstrate their technical athletic and qualitative artistic skills.
It is a key winter sport in many championships, including the Winter Olympic Games.
While many spectators enjoy watching the sport, it is often a mystery to understand how skater performances are judged.
Instead of being subjective or counterintuitive, judges actually use a formal process --- the International Skating Union (ISU) Judging System ~\cite{isu_rule} --- to score the performance of each skater.
Judges use a score sheet to inform and guide their scoring.
Formally, this score sheet is a \textit{rubric} with multiple criteria with separate rating scores that are additive.
For this work, we focus on the single-skating short-program event, where each skater performs for about three minutes, and is subjected to more rules and restrictions on their actions to execute, compared to free-form skating.
Table \ref{score_sheet} shows an example score sheet with judgement results from human judges during the 2014 European Figure Skating Championships. 

The total score consists of a combined Technical Element Score (TES) and a combined Program Component Score (PCS).
Skaters perform a sequence of elements (actions) in a schedule that is known ahead of the performance.
The chosen elements and how well each was executed affects the TES.
For example, a skater may choose \rev{a} sequence with easier elements, but would be expected to earn fewer points than another skater with a more complex sequence. This determines the \textit{Base} score.
For each element, the skater may execute it very well or poorly, receiving a positive or negative \textit{Grade of Execution (GOE)} score accordingly. GOE spans $-5$ to 5.
Each technical element is thus scored by adding Base and GOE.

There are three main types of elements --- Jumps, Spins, and Step Sequences --- and the Transition between them. 
A \textit{jump} is an action for which a skater leaps into the air, rotates at high speed, and lands after one or more revolutions. 
There are 6 types of jumps: Toe Loop (T), Salchow (S), Loop (Lo), Flip (F), Lutz (Lz), and Axel (A), which are distinguished by the way they take off and land when entering each jump.
Different types are variously difficult and will affect the expected score for the element.
Furthermore, each type of jump is assessed by the number of rotations; higher number of rotations are more difficult to execute. 
A \textit{spin} is a clockwise or counterclockwise rotation on ice with either foot on the rink. There are three basic types of spins: Upright (USp), Sit (SSp), and Camel (CSp). Spins can be performed individually or as a sequence combining different spin types with a change of position or foot, e.g., Change Foot Sit Spin (CSSp). For spins, judges see the difficulty, smoothness, and stability of movements. For example, a difficult entrance to the spin and continuous spins with stable posture will earn the skater a higher score.
A \textit{step sequence} is a technique that uses the entire rink for continuous footwork and performance. Skaters show steps and turns in a pattern, e.g., Rocker turn and Bracket turn, on the ice while fully utilizing the ice skating rink in accordance with the character of the music. There are three types of step sequences: Straight line step sequence (SISt), Circular step sequence (CiSt), and Serpentine step sequence (SeSt). Step sequences must match the music, be performed effortlessly throughout with good energy, flow and execution, and must have deep edges, clean turns and steps.

Other than scoring specific elements in the sequence, the skater is graded on overall qualities of her performance. This is indicated in the PCS score to judge various components: Skating Skills, Transition / Linking Footwork, Performance / Execution, Choreography / Composition, and Interpretation.
\textit{Skating Skills} score the overall cleanness, sureness, edge control, and flow over the ice surface demonstrated by a command of the skating vocabulary (edges, turns, steps, etc.), the clarity of technique, and the use of effortless power to accelerate and vary speed.
\textit{Transitions / Linking Footwork} scores the varied and purposeful use of intricate footwork, positions, movements, and holds linking all elements. Although this criterion may overlap with the TES for a step sequence, which include transitions and footwork, judges look at the entire performance to measure the quality to score this PCS component.
\textit{Performance / Execution} scores the involvement of the skater physically, emotionally, and intellectually as she delivers the intent of the music and composition.
\textit{Choreography / Composition} scores how intentionally developed and original is the arrangement of all types of movements according to the principles of the musical phrase, space, pattern, and structure.
\textit{Interpretation} scores the personal, creative, and genuine translation of the rhythm, character, and content of the music to movement.

\section{Technical approach}
We introduce IRIS, an \textbf{I}nterpretable \textbf{R}ubric-\textbf{I}nformed \textbf{S}egmentation method to predict and explain video-based action quality assessment (AQA).
It is \rev{\textit{ante-hoc}} interpretable~\cite{rudin2019stop, zhang2022towards} to explain how it derived its judgement score.
It is informed by a rubric to determine what to consider when calculating its judgement.
It performs segmentation to specify moments of a performance to judge specific criteria.
We implemented this for the application of figure skating, but the high-level approach is generalizable to other video-based AQA \rev{with established rubrics} (e.g., diving, dancing, cooking).

\begin{figure*}[th]
  \centering
  \includegraphics[width=11.0cm]{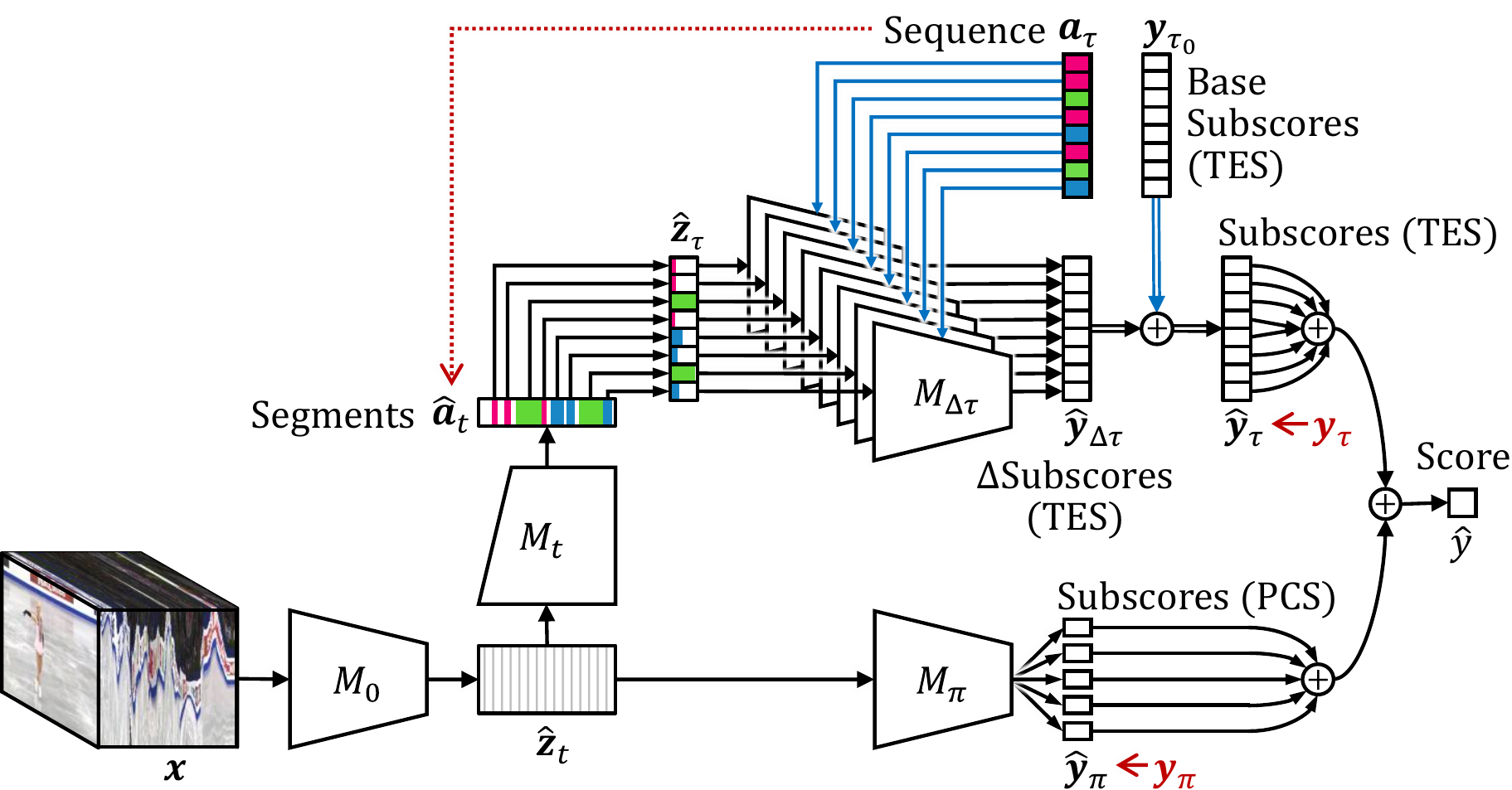}
  \caption{
  Architecture of IRIS showing how it leverages rubric information (Sequence $\hat{a}_t$ and base Subscores $\hat{\bm{y}}_{\tau_0}$) to provide explanations (Segments, TES GOE $\Delta$Subscores $\hat{\bm{y}}_{\Delta\tau}$, and TES $\hat{\bm{y}}_{\tau}$ and PCS $\hat{\bm{y}}_{\pi}$ subscores) to predict the Score $\hat{\bm{y}}$ of a figure skating performance in a video $\bm{x}$.
  Symmetrical trapeziods ($M_0$, $M_{\Delta\tau}$, $M_\pi$) represent convolutional neural networks, the half-rectangular trapezoid $M_t$ is a temporal convolutional network, and other rectangular shapes are variables (scalar, vector, tensors).
  Black and blue arrows indicate forward propagation, blue are specifically from rubric inputs, and red arrows indicate training loss.
  }
  \label{diagram_overview}
\end{figure*}

IRIS provides four key rubric features (Score, Sequence, Segments, and Subscores) to predict and explain its judgement.
Score is the final judgement, Sequence describes the actions to be judged (for technical elements), Segments identify when \rev{each} action was judged based on specific criteria, and Subscores indicate how each specific criteria was judged (for TES and PCS).
Fig. \ref{diagram_overview} illustrates the architecture of IRIS, in the following subsections.

\subsection{Data preparation}
Before modeling, we collected and processed data to gather necessary information for IRIS.
We obtained video data of figure skating from the MIT-Skate dataset~\cite{10.1007/978-3-319-10599-4_36} which includes 150 videos of short-event figure skating from various championships in 2008, 2010, 2012, and 2014. Each video is about three minutes long, which we converted to 4D tensors with 2D for x-y pixels, 1D for time, and 1D for color channels.
We collected rubric data as score sheets for each figure skater from online sources of the championship results maintained by ISU (e.g., ISU European Championships 2014\footnote{http://www.isuresults.com/results/ec2014/}). The score sheets are in PDF format, which we parsed with optical character recognition using Adobe Acrobat Pro v21.
From this, we obtained the skater name, TES base, GOE and total subscores, and PCS subscores and linked them to each skating performance video.
To train segmentation (described later), we manually annotated when each element was performed. This was done by researchers, with advice from practicing figure skaters.
With the data prepared, we can model the prediction for automatic judgment, which we describe next.
Finally, for simplicity and to provide sufficient data per class, we aggregated technical elements into broad categories of Jump, Spin, Step Sequence and Transition. Future work can train to recognize more specific action labels with more data.

\subsection{Base embedding for implicit knowledge}
The first step in IRIS is to predict a vector representation of the video. As is common for video data, we train a 3D CNN $M_0$, specifically I3D~\cite{i3d_arxiv}, to take a video tensor input $\bm{x}$ and predict an embedding $\hat{\bm{z}}_t$ of the video to represent information in the video.
It predicts a vector embedding for each \rev{0.534}-sec time window \rev{(16 frames at 29.97 fps like in \cite{cvpr_uncertainty,zeng2020hybrid})}. 
Each video has up to 356 windows with zero-padding, i.e., all videos are no longer than 3min 10s. 
Hence, $\hat{\bm{z}}_t$ is a 2D tensor with vectors across time.
Much prior work use these embeddings to predict the judgement score by inputting them into a feedfoward neural network~\cite{parmar_olympic}. We compare against this baseline in our modeling study later.

\subsection{Sequence of foreknown actions}
With IRIS, we extract sequence information $\bm{a}_\tau$ from the score sheet of each skater. This sequence of actions (technical elements) describes which actions to expect in sequence, but not specifically at what times. Nevertheless, assuming that the skater does not change her schedule, this sequence will be very informative to guide the overall scoring.
Indeed, human judges know ahead of time what actions to expect from the skater and can judge more accurately.
Furthermore, their judgement of each action leverages knowledge of whether the scheduled action is easy or difficult (affecting the base TES score), and helps the judge to narrow down their comparison of the action against similar ones (e.g., jumps against other jumps).
Surprisingly, this crucial information has been neglected in prior work, and our modeling study evaluation shows its strong effectiveness to improve score prediction performance.

Thus, we employed Sequence information for our predictions in IRIS.
Specifically, we used it to improve our Segmentation (described in the next section), and to improve TES subscore prediction.

\subsection{Segments of when actions happened}
We use the time series embeddings $\hat{\bm{z}}_t$ to predict the action $\hat{a}_t$ at each time window with a temporal convolutional network (TCN)~\cite{bai2018empirical}, specifically multi-stage TCN (MS-TCN)~\cite{farha2019ms} $M_t$. This is a sequence-to-sequence model that hierarchically convolves over time, considering short-term and long-term historical data. We trained $M_t$ using supervised learning with manually annotated segments as ground truth labels; e.g., we annotated when a jump starts and ends.

TCN prediction suffers from over-segmentation, where there are more segments predicted than the true segments; e.g., segment labels may flip-flop between different labels instead of being a continuous one.
First, as in~\cite{farha2019ms}, we used a smoothing loss regularization using the truncated mean squared error: 
$L_\mu = \frac{1}{T} \sum_t^T \max(\epsilon_t, \epsilon)$, where $\epsilon_t = (\log \hat{m}(t) - \log \hat{m}(t-1))^2$ is the squared of log differences and $\epsilon$ is the truncation hyperparameter. This merges small segments into fewer, larger ones.
Next, we employed a heuristic approach by 
i) counting the expected number of each action type $a$ known from the score card sequence information $n_a$, and 
ii) selecting the longest $n_a$ segments for each action type.
All remaining unselected segments were re-labeled as transitions.
This results in reasonable segmentation performance (see Fig. \ref{results_segments}).

With the time series action labels, we partition the time series embeddings $\hat{\bm{z}}_t$ based on when specific actions started and ended. These are the action sequence embeddings $\hat{\bm{z}}_\tau$. We zero-pad all embeddings to the same length for standardized processing next.

\subsection{Subscores of multiple specific criteria}
IRIS predicts several subscores, 7 TES subscore for separate technical elements, and 5 PCS subscores for the overall performance.
Having identified each sequence element, we can predict a TES subscore for each element.
Noting that the score cards used by human judges split TES subscores into Base and GOE, where Base is pre-determined, we realize that only GOE needs to be predicted.
Knowing the Base score is highly informative, thus predicting GOE is a much simpler problem which would be more accurate than predicting the full TES subscore.
Furthermore, knowing the expected element being performed at the sequence helps the judge to narrow the scope of judging.
Hence, for each sequence step $\tau$, we train a conditional convolutional neural network (c-CNN) $M_{\Delta\tau}$ that takes the sequence embedding $\hat{\bm{z}}_\tau$ with its corresponding actual action label $a_\tau$ to predict the GOE $\hat{y}_{\Delta\tau}$ for that sequence.
To obtain the total subscore for this element, we add the Base and GOE subscores, i.e., $\hat{y}_{\tau} = \hat{y}_{\tau_0} + \hat{y}_{\Delta\tau}$.
The total TES subscore for all technical elements are simply added together, i.e., $\hat{y}_{\Sigma\tau} = \sum_{\tau}{\hat{y}_{\tau}}$.

We predict PCS subscores differently, since they are based on overall performance instead of specific actions.
Using the time series embeddings of the whole video $\hat{\bm{z}}_t$, we train a multi-task CNN model $M_\pi$ to predict multiple PCS subscores $\hat{\bm{y}}_pi$ together.
Training a multi-task model instead of multiple, independent models improves accuracy due to correlations between the predictions.
The PCS subscores are combined into a total PCS subscore, i.e., $\hat{y}_{\Sigma\pi} = \sum_{\pi}{\hat{y}_{\pi}}$.

\subsection{Score}
Finally, the total Score is predicted by simply adding the TES and PCS scores together, i.e., $\hat{y} = \hat{y}_{\Sigma\tau} + \hat{y}_{\Sigma\pi}$.

In summary, IRIS exploits knowledge that judges already know from rubric score cards, and adheres to the criteria that human judges use to guide and account for their decisions.
These methods \rev{help} IRIS to be 
1) significantly more accurate, 
2) intrinsically interpretable to show its working to derive its final score, and
3) trustworthy since it uses the domain-standard rubric for judging.

\subsection{\rev{Implementation}}
\rev{We implemented IRIS using PyTorch and trained on an Nvidia RTX A6000 GPU, with a learning rate of 0.0005, batch size 60, and the Adam optimizer. 
Training converged within 300 epochs.
We trained IRIS to predict Segments $\hat{\bm{a}}_t$, TES $\Delta$Subscores $\hat{\bm{y}}_{\Delta\tau}$, PCS Subscores $\hat{\bm{y}}_{\pi}$ with uniform training loss hyperparameters, i.e., all 1.}

\begin{figure*}[h]
    \includegraphics[width=11.0cm]{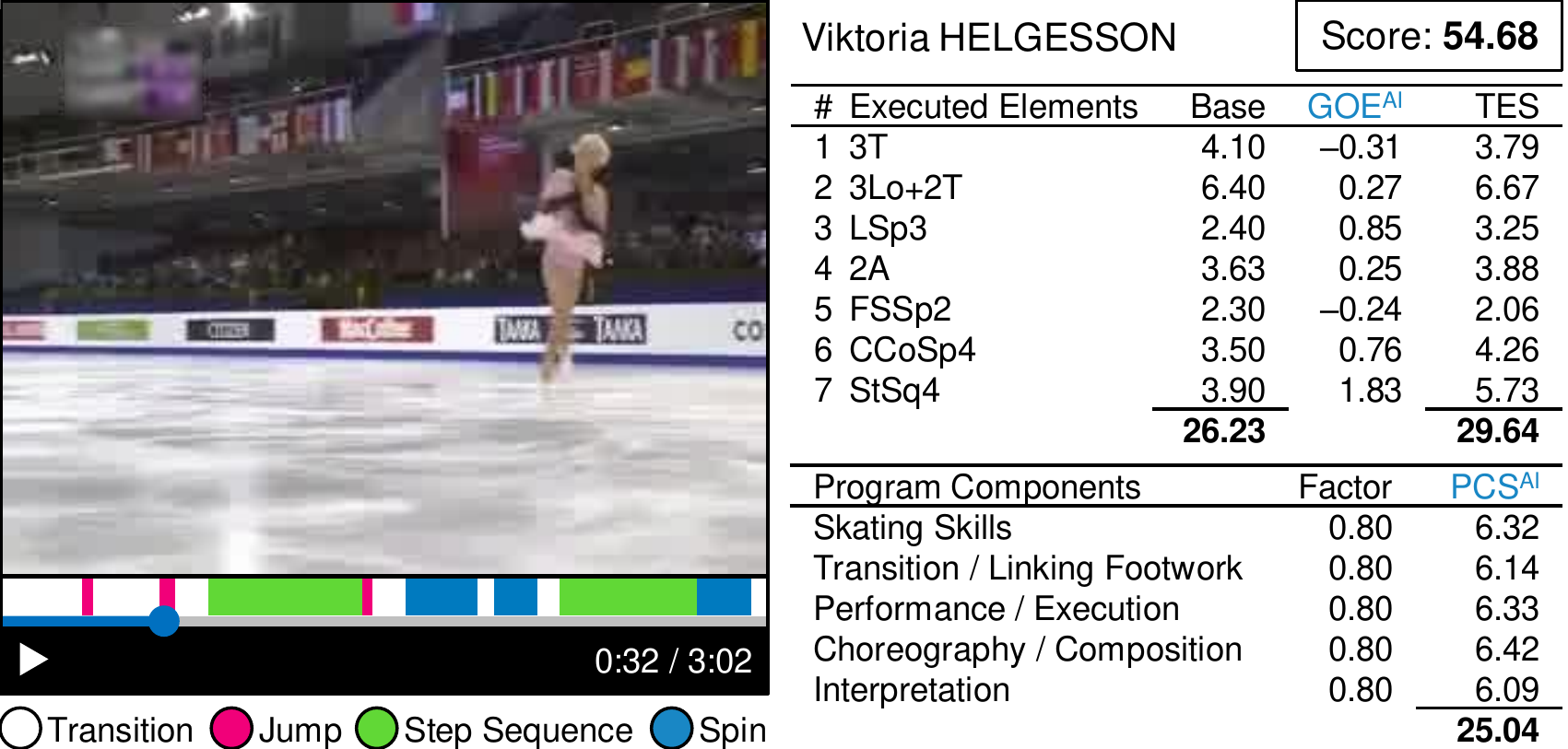}
    \caption{
    IRIS visualization for the UI variant \rev{showing the predicted judgement Score and} all rubric-informed explanations\rev{: Sequence, Segments, Subscores}.
    i) On the left, the video player with image frame, play button and current/end time is shown for all conditions, along with the Score number (top right).
    ii) The executed elements names and Sequence \# in the table is shown in the Sequence condition.
    iii) The colored timeline bar and legend of action colors is shown in the Segments condition.
    iv) The remaining GOE\textsuperscript{AI} and PCS\textsuperscript{AI} numbers in the score scheet table are shown in the Subscores condition.
    }
    \label{overview_ui}
\end{figure*}

\section{Evaluations}
We evaluated IRIS in three stages. 
First, we demonstrate its interpretability with a visualization of an ice skating performance.
Second, we conducted a quantitative modeling study comparing it against the baseline models.
Third, we conducted a formative user study with experienced figure skaters to qualitatively investigate the usefulness and trustworthiness of its rubric-informed features.

\subsection{Demonstration Study}

We presented the four key rubric-informed features in a unified visualization user interface (UI) shown to participants (see Fig. \ref{overview_ui}).
The key aspects of the user interface are a video player on the left, and score sheet on the right.
The total Score is displayed on the top-right corner.
To show Segments information, we augment the video player by overlaying \rev{a} color-annotated timeline over the seek bar. The colors indicate which of the four action types (transition, jump, spin, and step sequence) were automatically identified, and when they occurred. Users can examine specific actions by tracking the seek bar to the annotated location, and playing the video.

To present the Sequence and Subscores information, we leverage the score sheet layout typical of the ISU Judging System (IJS)~\cite{isu_rule}.
We simplified the table to minimally include the elements (to be) executed, Technical Elements Scores (TES), and Program Components Scores (PCS). 
The TES subscores are calculated for each element, and further subdivided into a Base score which is standard for the scheduled element, and known ahead of time, the AI predicted Grade of Execution (GOE) to judge how much the skater's performance was above or below average, and total TES subscore which is Base + GOE.
The PCS is also divided into separate program components (e.g., skating skills, choreography), which are determined across the whole skating performance. There is a factor multiplied on the PCS subscores, which is 1.00 for male skaters and 0.80 for female skaters.
The bold numbers at the bottom of columns indicate sums across elements or components.
Finally, experienced figure skaters know that the Score = TES + PCS.

\subsection{Modeling Studies}
We conducted two modeling studies on the performance of IRIS:
1) an ablation study to examine how each rubric-informed feature improves the model prediction performance, and
2) a \rev{benchmarking} study to evaluate whether IRIS performs better than baseline models.
Just like prior studies, all models were trained on randomly chosen 120 videos, and evaluated with a test set of 30 videos.
We describe the evaluation metrics, models compared, and results.

\begin{figure*}[h]
\includegraphics[width=\linewidth]{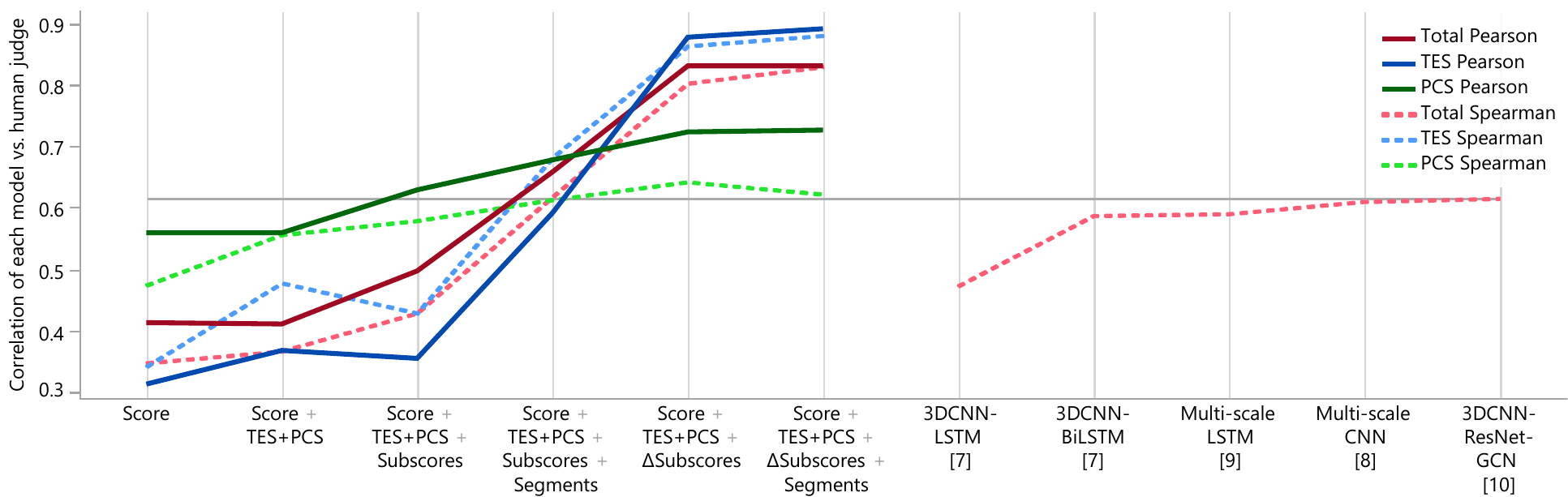}
\vspace{-0.5cm}
\caption{
\rev{Results of ablation study (Left) showing how predicting different interpretability features affects model performance in IRIS,
and model benchmarking study (Right) showing performance of competitive baselines.
Higher Spearman or Pearson correlation coefficients indicate better performance and more similarity between the model and human judges.
Horizontal gray line is highest benchmark performance for Spearman coefficient.
See Tables \ref{modeling_results_ablation} and \ref{modeling_results_comparison} for numeric details.}
}
\label{fig:modeling-results}
\end{figure*}

\subsubsection{Evaluation metrics}
For all comparisons, we evaluated how well each model can predict the human score in the dataset.
Most prior work evaluated using the Spearman rank correlation coefficient that indicates whether the ranking of skaters were preserved based on actual and predicted scores. However, this neglects how much the score may have changed. 
We further calculated the Pearson correlation coefficient to see how well IRIS can preserve the linearity in score predictions.
We also add to prior work by examining how well IRIS can predict TES and PCS subscores separately.

To evaluate segmentation performance, we calculated the Dice coefficient which indicates how much the predicted segment $\hat{\bm{a}}$ overlaps with the actual segment $\bm{a}$, i.e., $2(\bm{a} \cdot \hat{\bm{a}}) / (|\bm{a}|^2 + |\hat{\bm{a}}|^2)$.

\subsubsection{Comparison models}
For the ablation study, we evaluated different variants of the IRIS model to predict combinations of base and explanatory information:
Score $\hat{y}$, 
TES+PCS $(\hat{y}_{\Sigma\tau}, 
\hat{y}_{\Sigma\pi})$, 
Subscores $(\hat{\bm{y}}_\tau, \bm{y}_\pi)$,
$\Delta$Subscores $(\hat{\bm{y}}_{\Delta\tau}, \bm{y}_\pi)$,
Segments $\hat{\bm{a}}_t$.

For the \rev{benchmarking} study, we evaluated IRIS against the following state-of-the-art models:
\aptLtoX[graphic=no,type=env]{
\begin{enumerate}[leftmargin=*]
    \item[1)] 3DCNN-LSTM~\cite{parmar_olympic} puts video embeddings into an LSTM to predict the total scores with long-duration patterns.
    \item[2)] 3DCNN-BiLSTM~\cite{parmar_olympic} uses Bi-directional LSTM instead of plain LSTM to capture time patterns forwards and backwards in the video. This reduces early forgetting. 
    \item[3)] 3DCNN-ResNet-GCN~\cite{zeng2020hybrid} uses a video 3DCNN model with graph convolutional network to model cross-dependencies between video segments.
    \item[4)] Multi-scale CNN~\cite{nekoui2021eagle} uses pose heatmaps, and varying CNN kernel sizes to capture short and long time dependencies.
    \item[5)] Multi-scale LSTM~\cite{8756030} aggregates multiple LSTM models to capture temporal patterns at varying time resolutions.
\end{enumerate}
}{
\begin{enumerate}[label={\arabic*)}, leftmargin=*]
    \item 3DCNN-LSTM~\cite{parmar_olympic} puts video embeddings into an LSTM to predict the total scores with long-duration patterns.
    \item 3DCNN-BiLSTM~\cite{parmar_olympic} uses Bi-directional LSTM instead of plain LSTM to capture time patterns forwards and backwards in the video. This reduces early forgetting. 
    \item 3DCNN-ResNet-GCN~\cite{zeng2020hybrid} uses a video 3DCNN model with graph convolutional network to model cross-dependencies between video segments.
    \item Multi-scale CNN~\cite{nekoui2021eagle} uses pose heatmaps, and varying CNN kernel sizes to capture short and long time dependencies.
    \item Multi-scale LSTM~\cite{8756030} aggregates multiple LSTM models to capture temporal patterns at varying time resolutions.
\end{enumerate}
}

\subsubsection{Results}
\begin{figure}[t]
\vspace{-0.1cm}
\includegraphics[width=8.5cm]{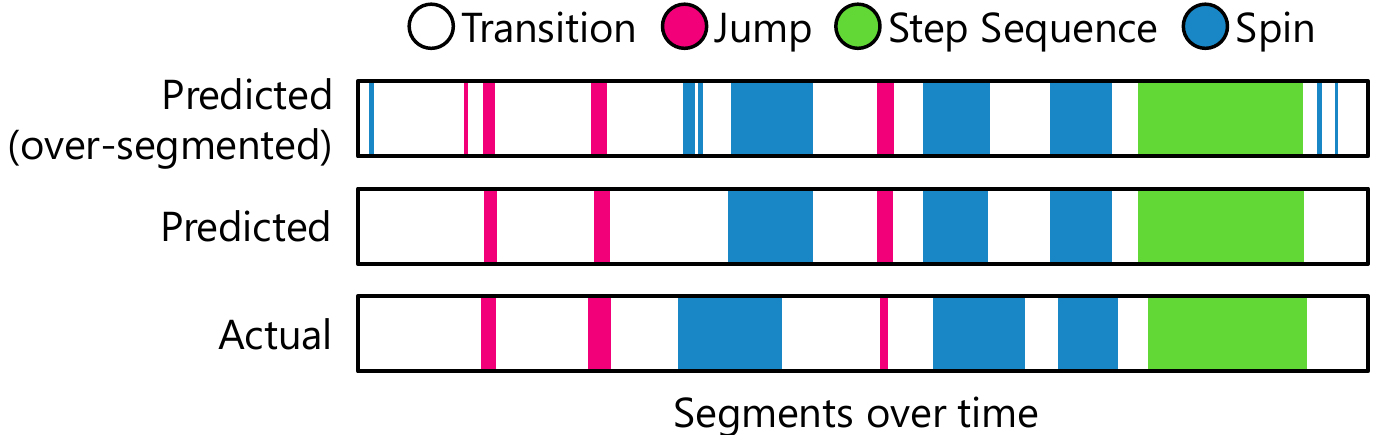}
\vspace{-0.4cm}
\caption{
Example predicted and actual segments for a skating sequence showing:
1) the initial prediction with over-segmentation,
2) corrected predicted segmentation, which matches reasonably closely with
3) the actual segmentation.
}
\vspace{-0.3cm}
\label{results_segments}
\end{figure}

\begin{figure}[t]
  \centering
  \includegraphics[width=4.6cm]{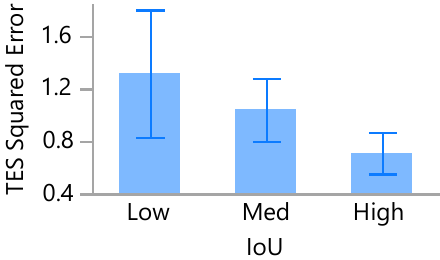}
  \vspace{-0.1cm}
  \caption{
  \rev{Error in TES prediction is lower with better segmentation IoU.
  IoU divided by tertiles Low (0-33\%-tile), Med (33-66\%), High (66-100\%).
  Error bars indicate standard error.}
  }
  \label{tes_iou_results}
\end{figure}

Fig. \ref{fig:modeling-results} (Left) and Appendix Table \ref{modeling_results_ablation} shows the results from our ablation study.
As expected, combining all interpretability features led to the best performing model.
Predicting with subscores helped IRIS to improve its total score prediction performance, indicating that decomposed analysis was beneficial.
Leveraging the TES base subscores helped to significantly improve the performance for predicting TES.
Segmentation helped to improve TES subscore prediction performance significantly, TES $\Delta$subscore prediction performance slightly, but not necessarily for PCS.
The segmentation in IRIS had good a Dice coefficient of 0.79. Fig. \ref{results_segments} shows an example segmentation. \rev{Better segmentation with higher IoU also generally resulted in better TES score prediction as shown in Fig. \ref{tes_iou_results}.}

Fig. \ref{fig:modeling-results} (Right) and Appendix Table \ref{modeling_results_comparison} shows the results from our \rev{benchmarking} study.
The basic 3DCNN and multi-task predictions in simpler IRIS models were weaker than other baselines, but IRIS was better when it modeled with TES base subscores and segmentation.
This indicates that being informed with rubrics and including interpretability can be more beneficial than using esoteric, generic time-series features that other baselines have.

\subsection{Formative User Study}

Having shown that IRIS improves model prediction performance, we next evaluated whether the explanations provided by IRIS \rev{are} useful.
We did so with practicing figure skaters who have domain expertise to meaningfully interpret the explanations.
Our research objective is to determine the \textit{usefulness} and \textit{trustworthiness} of various rubric-informed explanations for automatic action quality assessment in figure skating videos.
We investigated why users found each explanation feature useful and how they used the feature.
The study was approved by our institution's institutional review board.

\subsubsection{Experiment Method}
We conducted a formative user study to collect user opinions and experiences on different features of IRIS by implementing different variants of the IRIS visualization.
Each variant progressively adds more UI features:
\begin{itemize}[leftmargin=*]
  \item \textbf{Score (S)} that shows the AI prediction to judge the performance.
  \item\textbf{S + Sequence (SS)} that also shows the sequence of technical elements that the performer planned to execute. This is what human judges know ahead of each judging session. It does not include any subscore information for each technical element.
  \item \textbf{SS + Segments (SSS)} that also shows the AI prediction of the time segments when each technical element was executed. 
  \item \textbf{SSS + Subscores (SSSS)} that also shows all TES  (base and GOE) and PCS subscores. 
  This is the most comprehensive UI.
\end{itemize}

For simplicity, we implemented the UI variants in presentation slides, though the data is computed by the IRIS model, and thus realistic of the model behavior.
We presented \rep{all variants within-subjects} to the participant, with fixed order: S, SS, SSS, SSSS. 
This was not counterbalanced, since each latter variant contains more information and would confound with a learning effect.
We prioritized a qualitative formative study instead of a larger-scale summative study to gather nuanced insights regarding the use of rubric-informed explanations, and because of the difficulty to recruit a large number of trained domain experts (practicing figure skaters).

\subsubsection{Experiment Procedure}
The procedure each participant was:
\aptLtoX[graphic=no,type=env]{
\begin{enumerate}[leftmargin=*, topsep=0pt]
    \item[1)] \textit{Introduction} about the experiment objective and procedure, and description about 
    the figure skating videos.
    \item[2)] \textit{Consent} to participate with their voice and interactions recorded.
    \item[3)] Two video sessions with:
    \begin{enumerate}
        \item[i)] \textit{Video} is played lasting about 3 minutes.
        \item[ii)] \textit{Score (S)} prediction is displayed after the video is played, followed by several interview questions discussed (details later).
        \item[iii)] \textit{Sequence (SS)} information of the technical elements is next displayed and questions posed again.
        \item[iv)] \textit{Segments (SSS)} of when each action was performed are next displayed and questions posed again.
        \item[v)] \textit{Subscores (SSSS)} are displayed and questions posed again.
        \item[vi)] \textit{Interview questions} to ask about:
        \begin{enumerate}
            \item[1)] Do you agree or disagree with the AI judgement score? [7-pt Likert]. Why or why not?
            \item[2)] How do you think the AI scored the performance? Do you agree or disagree with its method? Why or why not?
            \item[3)] Do you agree or disagree that you understand how AI predicts scores? (Perceived Understanding) [7-pt Likert]
            \item[4)] Would you score differently? How so?
            \item[5)] Which part of the UI did you find (not) useful? Why (not)?
            \item[6)] Any suggestions for improvement?
        \end{enumerate}
    \end{enumerate}
\end{enumerate}
}{
\begin{enumerate}[label={\arabic*)}, leftmargin=*, topsep=0pt]
    \item \textit{Introduction} about the experiment objective and procedure, and description about 
    the figure skating videos.
    \item \textit{Consent} to participate with their voice and interactions recorded.
    \item Two video sessions with:
    \begin{enumerate}[label={\roman*)}]
        \item \textit{Video} is played lasting about 3 minutes.
        \item \textit{Score (S)} prediction is displayed after the video is played, followed by several interview questions discussed (details later).
        \item \textit{Sequence (SS)} information of the technical elements is next displayed and questions posed again.
        \item \textit{Segments (SSS)} of when each action was performed are next displayed and questions posed again.
        \item \textit{Subscores (SSSS)} are displayed and questions posed again.
        \item \textit{Interview questions} to ask about:
        \begin{enumerate}[label={\arabic*)}]
            \item Do you agree or disagree with the AI judgement score? [7-pt Likert]. Why or why not?
            \item How do you think the AI scored the performance? Do you agree or disagree with its method? Why or why not?
            \item Do you agree or disagree that you understand how AI predicts scores? (Perceived Understanding) [7-pt Likert]
            \item Would you score differently? How so?
            \item Which part of the UI did you find (not) useful? Why (not)?
            \item Any suggestions for improvement?
        \end{enumerate}
    \end{enumerate}
\end{enumerate}
}

\subsubsection{Findings}

We recruited 10 participants through snowball sampling from two university figure skating clubs and one organization in Japan. 
Discussions were conducted in Japanese and translated to English.
\rev{We acknowledge the sample limitation due to the challenge of recruiting rare expertise, yet benefited from very interesting expert feedback.}
Their average age was 23.8 (SD = 8.6) years old and 8 were female.
All had experience in figure skating spanning 2-13 years (M = 6.6).
We determined participant expertise by asking their Skill Test Level~\cite{isu_skill}. 
One participant had not taken the test, thus we considered her a novice. 
Others were skilled with M = 2.6 out of 5 (SD = 1.7), and max = 5. 
%
Due to Covid-19 restrictions, the experiment was conducted remotely over Zoom with screen sharing and audio recording.
The interview sessions took 81.8min on average (SD = 10.5) with \$20 USD compensation.

\begin{figure*}[h]
  \centering
  \includegraphics[width=\linewidth]{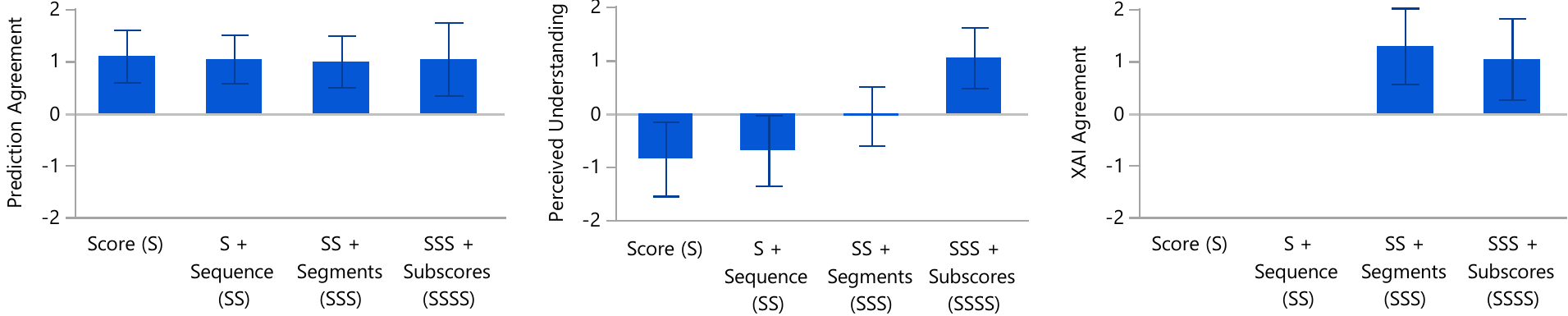}
  \vspace{-0.1cm}
  \caption{
  Mean participant ratings on agreement with AI score, perceived understanding of AI judgement, and agreement with UI Variants with different XAI features (S, SS, SSS, SSSS).
  Error bars indicate 95\% confidence interval.
  }
  \label{user_study_2nd_plot}
\end{figure*}

Figure \ref{user_study_2nd_plot} shows the quantitative results of participant ratings.
Participants somewhat agreed with the AI prediction score (M = 1.05), though there was no difference among UI variants.
They found the full UI variant with all explanations easiest to understand (M\textsubscript{SSSS} = 1.05), followed by the UI with Segments (M\textsubscript{SSS} = $-$0.05), but did not understand how the AI worked when only viewing the Score (M\textsubscript{S} = $-$0.85) and Sequence variants (M\textsubscript{SS} = $-$0.70).

Next, we analyzed the recorded sessions and performed a thematic analysis in terms of our research objective to understand the helpfulness and trustworthiness of each rubric-informed feature. 
Themes were generated and refined by the first and last authors.

\textbf{Score only.}
Participants mostly depended on their own experience and intuition to judge how the score was derived, or were clueless about \rev{how the} AI could perceive and make judgements.
P4 mentioned that she \textit{"didn't know and didn't have a clue"}.
P5 saw that the skater \textit{"was off-balance on the first jump, so [he] was going to drop [the score] a little more [than what the AI scored]"}.
P10 \textit{"felt that it was difficult to understand the scoring method from the images and scores alone, and felt that it would be most convincing if we could come up with a protocol like the scoring table used in competitions"}.
Some participants attributed advanced capabilities \rev{to} the AI, even though this was not implemented. P2 \textit{"thought the AI was scoring the number of jump rotations, the flow after landing, and so on"}.

\textbf{Sequence.}
With the sequence information, many participants realized that the AI considered specific actions in its judgement.
P2 remarked that \textit{"by  knowing  the  elements  in  advance,  I  knew  approximately  how  many points the basic score would be, which made it easier to predict the score, and I thought [the AI's score] was reasonable."}.
P3 noted that he had interpreted the AI reasoning in the Score condition \textit{"by guessing earlier, but had suspected that [the AI] was calculating based on the basic points of each element."}
The Sequence enabled participants to discuss the scoring more carefully. P5 saw that the skater \textit{"was doing 3T+3T with low score for a triple-turn combo, but I thought [the score] could have be a little higher."}

\textbf{Segments.}
Participants highly appreciated seeing the action segmentation explanations from the AI. This allowed them to perceive what the AI could notice.
P8 \textit{"was amazed that the steps were properly recognized. I thought that if it could recognize them, it must be able to judge the movements of the feet as well."}
Several participants felt that the segments would be helpful for less-knowledgeable spectators.
\textit{"Jumps and spins can be understood even by people who don't know skating, but some people don't know about steps, so it may be useful for them"} (P10).
P2 \textit{"was impressed to see that the jumps, spins, and steps were detected"}. 
However, he also remarked \textit{"on the other hand, it is difficult to judge whether the score is appropriate 
because it is not clear at what step and for what reason the score is given."}
Thus, he and other participants still wondered how each segment was scored.
Some participants also noticed inaccuracies in the segmentation.
P5 \textit{"thought the spin detection was a little fast. It appeared that the step had started before the step sequence was detected."}
P6 noted that \textit{"the end of spin detection was early."}
Nevertheless, these participants remained positive about the AI segmentation capability.

\textbf{Subscores.}
With the subscores, participants found the AI most useful.
P3 felt that \textit{"it is easier to be convinced if you can see the detailed scores. I thought it was more convincing when I could see the completion of each element (e.g., first jump, fifth spin, etc.), and the points for possible mistakes were minus."}.
P9 felt that the AI \textit{"can be evaluated objectively without human error. It is useful to visualize how much points are added or subtracted from the base score."}
The subscores stimulated much discussion about the judging correctness.
P1 argued that \textit{"fewer points should have been deducted for [the] 3Lo [technical element], but everything else seems to be OK."}
P2 remarked that \textit{"3T3T looked very well put together, so more points should be awarded. I thought FCSp3 deserved a higher score because [the skater] was doing something more difficult than the usual Camel spin. The (high) score for Choreography was surprisingly satisfactory.}"

Participants also requested for more details.
P3 wanted to \textit{"know the type of jump."}, rather than just knowing that one was observed.
\textit{"It would be nice to know the types of spins, postures, deformations, as well as the types and difficulties of spins. This would be useful for judging the level of the skaters"} (P3)\rev{.}
P9 \textit{"I would also like to know the height of the spin, and whether 3D expression is achieved."}

Participants were mostly positive about the TES subscore predictions, but had varying opinions about the more qualitative PCS subscores.
P5 identified that \textit{"the PCS scores, such as Skating skills, were all in the 6-point range. The top scorers get about 7 points, so I thought [the AI's score] was reasonable."}
P7 found \textit{"the difference in PCS scores is disconcerting. I don't think it would be like this if a person gave it."}
The nuance and subjectivity in judging PCS also raised questions.
P1 commented that \textit{"even if [the PCS score] is decomposed, I wonder how it is determined."}
Participants also needed more information to help to properly assess some PCS aspects.
P8 found that \textit{"it is hard to know how the expressiveness of the music, which is one of the best aspects of figure skating, was scored."}

\textbf{In summary}, participants found that the latter explanation features aligned well with their expected criteria for judging figure skating, and highly appreciated the increasing details to account for the final score. 
The potential of IRIS is neatly expressed by P9: 
\textit{"I think that the introduction of AI will be beneficial to the players because it will eliminate bias and subjectivity in this area, as I believe that human preferences have various influences."}

\section{Discussion}
We have shown that IRIS performed better than the current methods to automatically judge figure skating, and is perceived as more understandable and trustworthy by domain experts. 
We discuss limitations and generalizations of IRIS for interpretable AQA.


\subsection{Generalizing rubric-informed AQA}
In this work, we studied figure skating as a representative example of sports that are qualitatively judged, yet have established rubrics for consistency. 
\rev{We believe that IRIS can be used to automatically score other qualitative sports or performances that are guided by rubrics, such as, dancing, skiing, diving.}
Note that while some sports activities are very brief, they can still contain a sequence of actions.
For example, a diving action consists of jumping from a diving board, performing an acrobatic sequence in the air, and entering the water. Appropriate labeling of these steps would provide more detailed information to divers and audiences and improve their understanding of the performance. 
\rev{IRIS is not meant for sports that use other scoring methods, such as counting goals (e.g., soccer, basketball, golf, fencing, boxing).
IRIS is not designed to generate a rubric; decision tree or rule mining approaches can already do this, albeit with the risk of learning spurious relationships.}

\subsection{Need for interpretability of artistic quality}
In this work, we provided detailed information on the technical and artistic criteria of figure skating \rev{by revealing additive segments and subscores}. 
In our user study, our participants requested for \rev{more justification on how each technical element or overall qualities are considered high or low scoring.}
\rev{Future work can provide} two approaches to address this.
1) Use XAI saliency map techniques (e.g., \cite{selvaraju2017grad}) to highlight important pixels or poses that the model focused on for predicting each TES or PCS score. However, without properly constraints (e.g., on pose joints), the highlights may be spurious.
2) Use heuristic methods based on meaningful movement features, such as Laban Movement Analysis (LMA)~\cite{bartenieff2013body} that interprets the artistic movement with four dimensions (Body, Effort, Shape, and Space) to describe the quality of movement, body posture and shape, and how the body is using the space. A good quality as defined with LMA should correspond to a high technical or artistic score.

\section{Conclusion}

We proposed IRIS, an \textbf{I}nterpretable \textbf{R}ubric-\textbf{I}nformed \textbf{S}egmentation method to predict and explain video-based action quality assessment (AQA). 
IRIS is:
1) interpretable to account for its judgement process,
2) informed by a rubric to judge by specific criteria, and 
3) performs segmentation to locate specific sections to judge each criteria.
We implemented IRIS for the application domain of figure skating to predict a Score and explain with Sequence, Segments, and Subscores.
As a result, by being rubric-informed, IRIS is better guided to make more accurate score predictions, is more accountable by justifying its decisions based on domain-accepted criteria, and hence more trustworthy to practicing figure skaters.

\begin{acks}
This work was supported by 
the Singapore Ministry of Education (MOE) Academic Research Fund Tier 2 T2EP20121-0040, and
the Japan Science and Technology Agency (JST) CREST Program JPMJCR21F2.
\end{acks}

\bibliographystyle{ACM-Reference-Format}
\bibliography{manuscript}

\clearpage

\appendix

\onecolumn
\section{Appendix}

\renewcommand{\arraystretch}{1.5}
\begin{table*}[h]
\caption{
Results of ablation study showing how predicting different interpretability features affects model performance in IRIS.
Higher Spearman or Pearson correlation coefficients indicate better performance.
\textbf{Bold} indicates best performance.
See Fig. \ref{fig:modeling-results}.
}
\label{modeling_results_ablation}
\begin{center}
\begin{tabular}{clcccclccclccc}
\hline
\multicolumn{6}{c}{\multirow{2}{*}{Features}} & \multicolumn{1}{c}{} & \multicolumn{7}{c}{Metrics} \\ \cline{8-14} 
\multicolumn{6}{c}{} & \multicolumn{1}{c}{} & \multicolumn{3}{c}{Spearman} &  & \multicolumn{3}{c}{Pearson} \\ \cline{1-6} \cline{8-10} \cline{12-14} 
\multicolumn{2}{c}{\renewcommand{\arraystretch}{1.0}\begin{tabular}[c]{@{}c@{}}Score \\ $\hat{y}$\end{tabular}} & \renewcommand{\arraystretch}{1.0}\begin{tabular}[c]{@{}c@{}}TES+PCS \\ $(\hat{y}_{\Sigma\tau}, \hat{y}_{\Sigma\pi})$\end{tabular} & \renewcommand{\arraystretch}{1.0}\begin{tabular}[c]{@{}c@{}}Subscores \\ $(\hat{\bm{y}}_\tau, (\hat{\bm{y}}_\pi)$\end{tabular} & \renewcommand{\arraystretch}{1.0}\begin{tabular}[c]{@{}c@{}}$\Delta$Subscores \\ $(\hat{\bm{y}}_{\Delta\tau}, (\hat{\bm{y}}_\pi)$\end{tabular} & \renewcommand{\arraystretch}{1.0}\begin{tabular}[c]{@{}c@{}}Segments \\ $\hat{\bm{a}}_t$\end{tabular} &  & \multicolumn{1}{c}{TES} & \multicolumn{1}{c}{PCS} & \multicolumn{1}{c}{Total} &  & \multicolumn{1}{c}{TES} & \multicolumn{1}{c}{PCS} & \multicolumn{1}{c}{Total} \\ \cline{1-6} \cline{8-10} \cline{12-14} 
\multicolumn{2}{c}{\checkmark} &  &  &  &  &  & 0.341 & 0.473 & 0.347 &  & 0.314 & 0.560 & 0.414 \\
\multicolumn{2}{c}{\checkmark} & \checkmark &  &  &  &  & 0.477 & 0.555 & 0.366 &  & 0.369 & 0.560 & 0.412 \\
\multicolumn{2}{c}{\checkmark} & \checkmark & \checkmark &  &  &  & 0.428 & 0.578 & 0.428 &  & 0.356 & 0.629 & 0.498 \\
\multicolumn{2}{c}{\checkmark} & \checkmark & \checkmark &  & \checkmark &  & 0.681 & 0.612 & 0.617 &  & 0.592 & 0.678 & 0.659 \\
\multicolumn{2}{c}{\checkmark} & \checkmark &  & \checkmark &  &  & 0.863 & \textbf{0.641} & 0.802 &  & 0.877 & 0.723 & \textbf{0.831} \\
\multicolumn{2}{c}{\checkmark} & \checkmark &  & \checkmark & \checkmark &  & \textbf{0.880} & 0.621 & \textbf{0.829} &  & \textbf{0.891} & \textbf{0.726} & \textbf{0.831} \\ \hline
\end{tabular}
\end{center}
\end{table*}

\renewcommand{\arraystretch}{1.5}
\begin{table*}[h]
\caption{
Spearman correlation coefficients of the total Score prediction for different baseline models.
Numbers are quoted from the original papers.
Refer to Table \ref{modeling_results_ablation} to compare different IRIS model variants against these baselines.
See Fig. \ref{fig:modeling-results}.
}
\label{modeling_results_comparison}
\begin{center}
\begin{tabular}{lr}
\hline
Models & Spearman \\ \hline
3DCNN-LSTM~\cite{parmar_olympic} & 0.479 \\
3DCNN-BiLSTM~\cite{parmar_olympic} & 0.587 \\
3DCNN-ResNet-GCN~\cite{zeng2020hybrid} & 0.615 \\
Multi-scale CNN~\cite{nekoui2021eagle} & 0.610 \\
Multi-scale LSTM~\cite{8756030} & 0.590 \\ \hline
\end{tabular}
\end{center}
\end{table*}

\clearpage

\end{document}